\documentclass{article}
\usepackage{aaai}
\usepackage{theorem}
\begin{document}

\title{Optimal Belief Revision}
\author{Carmen Vodislav  \hspace{1cm} Robert E. Mercer \\[5mm]
Cognitive Engineering Laboratory \\
Department of Computer Science \\
The University of Western Ontario \\
London, Ontario, Canada N6A 5B7 \\[2.5mm]
\{vodislav, mercer\}@csd.uwo.ca}

\maketitle

\begin{abstract}

We propose a new approach to belief revision that provides a way to
change knowledge bases with a minimum of effort. We call this way of
revising belief states optimal belief revision. Our revision method
gives special attention to the fact that most belief revision processes
are directed to a specific informational objective.
This approach to belief change is founded on notions such as optimal
context and accessibility. For the sentential model of belief states we
provide both a formal description of contexts as sub-theories determined
by three parameters and a method to construct contexts.  Next, we
introduce an accessibility ordering for belief sets, which we then use
for selecting the best (optimal) contexts with respect to the processing
effort involved in the revision. Then, for finitely axiomatizable
knowledge bases, we characterize a finite accessibility ranking from
which the accessibility ordering for the entire base is generated and
show how to determine the ranking of an arbitrary sentence in the
language. Finally, we define the adjustment of the accessibility ranking
of a revised base of a belief set.
\end{abstract}

\section{Introduction}
In this paper we put forward a new approach to changes in knowledge
bases. Two main reasons led us to this approach to belief revision.
One of these reasons originates in the intuition that in a belief
revision process only part of the knowledge base has an active role in
the integration of a new belief in the base. Therefore a belief revision
process involving only that sub-system of beliefs can capture the
essence of this process and reach the same effects as the traditional
way of belief changing with a diminished processing effort.
The other idea comes naturally when the belief revision process is
placed in a communication environment.\footnote{This approach was
suggested by Sperber and Wilson's relevance theory.
Refer to Sperber and Wilson (1986).} People are changing their minds
when interacting with the world, including the others' beliefs, and
usually when they are looking for answers to their questions.  Not every
fact or evidence is worth incorporating in our belief systems but only
information that is relevant for us with respect to our informational
needs. Consider that two intelligent
agents are involved in a goal-driven dialogue. One of them, called the
inquirer, is ``interested" in acquiring a piece of information that would
produce, by means of a belief revision process, the fulfillment of its
objective or {\em desideratum}. The other agent's role is to present {\em facts}
that would help the inquirer attain its goal.

Given an incoming piece of information, one's beliefs, and the
desideratum, our purpose is to process the available
information {\em efficiently} in order to fulfill the desideratum. We use
terms like {\em optimal} and {\em efficiency} throughout this paper with the meaning
of a comparison or ratio between the effects and effort corresponding to
a belief revision process. However, these terms are not related to
computational complexity. Instead, the {\em effects} represent
information derived when incorporating a piece of evidence in the
knowledge base; the {\em effort} required to reach such effects is mostly
characterized by the notion of {\em accessibility}. A formal definition of the
accessibility ordering will be provided later in this paper.

We call this way of revising belief systems {\em optimal belief revision}.
Defining a way of revising a knowledge base efficiently reduces to
choosing {\em optimal contexts}. The notion of context is related to and
determined by three parameters: the new belief, one's knowledge base and
the desideratum, and it represents any part of the knowledge base that
actually contributes to the integration of the new belief. The optimal
context is a context in which the new piece of information will be the
most efficiently processed such that the desideratum is met.

Both constructive and descriptive approaches to contexts are considered.
In the third section we stipulate the conditions that circumscribe the
set of context candidates, from which the optimal context is chosen. In
the fourth section we give an explicit construction for (optimal) contexts.
In the fifth section we introduce the concept of accessibility.
Finally, for finitely axiomatizable knowledge bases
(theories\footnote{ By theory we mean
any set of sentences closed under the consequence
operation {\it Cn} provided by the classical propositional calculus.}), we
define a finite accessibility ranking from which the accessibility
ordering for the entire base is generated. Also, we will see how from a
finite accessibility ranking one can adjust the ranking of a repeatedly
revised belief set.

\section{Related Work}
We mention three important sources related to our work. Firstly, the
axiomatic and constructive approaches for contraction functions
presented in G\"{a}rdendors (1988) serve as models for characterizing (optimal)
contexts from the same perspectives. Moreover, the definition and
analysis of epistemic entrenchment in relation to contraction functions
guided us in defining the accessibility notion and relate it to
modelling optimal contexts.
Secondly, relevance theory introduced in Sperber and Wilson (1986)
motivated us to
apply relevance theory to belief revision by placing agents in a
communicative environment.
Thirdly, in Williams (1998), Mary-Anne Williams uses a finite partial
entrenchment ranking to generate an entrenchment ordering and describes
a computational model that adjusts this ranking using an absolute
measure of minimal change. Later sections of our paper focus on the
same matters using the accessibility relation instead.

\section{Contexts: Descriptive Approach}
Consider a language {\bf L} whose logic includes classical propositional
logic. Beliefs and contexts are represented by closed sets of sentences
(belief sets).

Let $K \subseteq \mathbf{L}$ be the initial belief set of the inquirer. Suppose that the
inquirer's objective/desideratum is based on a {\em presupposition} $P$, which
is a sentence in {\bf L}, precisely a disjunction of possible answers to a
question: P = $G_1 \vee G_2 \vee \ldots \vee G_n$.\footnote{The notions of desideratum and presupposition
originate in Hintikka's approach to interrogatives; see Hintikka (1989), p.158.}
We consider that the presupposition is one of the
beliefs in K, $P \in K$.

Each  $G_i \in \mathbf{L}$ is called a basic goal. A disjunction of at most $n-1$ such basic
goals is considered also a goal. Any goal achieves the desideratum, but
a basic goal is preferred rather than a goal.

To attain its desideratum,
the inquirer must derive one of these goals from a belief revision
process. Let $A \in \mathbf{L}$ be the new piece of information
acquired by the inquirer, which is used in the belief revision process.

In this setting, contexts are considered subsets of $K$ that are also
closed under logical consequence.
Choosing a context whenever some new evidence is incorporated in $K$
depends upon three parameters: the belief set $K$, the new evidence $A$ and
a certain goal $G$ or desideratum. They will appear as constants
throughout our analysis. Contexts are considered optimal if for some $K$,
$A$ and $G$, the effort required for processing $A$ in that context such that
$G$ is derived is as small as possible.  Both the notion of accessibility
and the number of sentences in a context have to be considered in
assessing a certain processing effort.

The next theorem\footnote{The proofs of all the theorems can be found
in Vodislav (2000).} provides a description of (potentially optimal)
contexts given the above-cited parameters. The first part ({\it i)}) of
Theorem 1 characterizes contexts when no desideratum or goal is assumed.
The second part adds new conditions for contexts when this parameter is
considered as well. It is supposed that the revision, expansion and
contraction operators respect the AGM postulates.

\begin{theorem}
Let $K$ be a belief set and ``*" a revision operator. Let us
denote by ``$-$" the contraction operator that is related to ``*" through
the Levi Identity.\footnote{\ldots, that is, for any belief set $K$, $K^*_A =
(K^-_{\neg A})^+_A$.
See G\"{a}rdenfors (1988) p.69.}
\begin{enumerate}
\item [{\it i)}] For any belief set $K^\prime \subseteq K$, if 
      \begin{equation} K^\prime \setminus {K^\prime}^-_{\neg A} = K \setminus K^-_{\neg A},
      \mbox{\ then}
      \end{equation}
      \begin{equation} K^*_A = \mathit{Cn}[{K^\prime}^*_A \cup (K \setminus K^\prime)].
      \end{equation}
\item [{\it ii)}] If there is a sentence $G \in \mathbf{L}$, such that \\
(a) $G \notin K$;
(b) $G \in K^*_A$ \\ then there is a belief
set $K^\prime \subseteq K$ satisfying condition (1) and:
\end{enumerate}
\begin{itemize}
\item ${K^\prime}^-_{\neg A} \neq \emptyset$ if $A \not\vdash G$;
\item $G \in {K^\prime}^*_A$;
\item If $\neg G \in K$ then $\neg G \in K^\prime$.
\end{itemize}
\end{theorem}

Since in the limiting case $K^\prime$ can be identified with $K$, the first part of
Theorem 1 says that for any belief set $K$ we can always identify a
sub-theory $K^\prime$ of $K$, whose revision with a new belief $A$ tells us exactly
what sentences are retracted from $K$ when revising $K$ with $A$.
$K^\prime$ is chosen
such that exactly the same beliefs are retracted from both
$K$ and $K^\prime$ when revising these belief sets with the same
sentence $A$. This means that no sentences that could contribute to the
derivation of $A$ are left in the remainder $K \setminus K^\prime$.
Moreover, attaching the revised ``context"  to this remainder does not
lead to the negation of $A$, unless $\neg A$ is a tautology. The
relationship between a belief set $K$ and one of its contexts
associated to $A$ are presented in the following corollary:

\begin{corollary}
For a revision operator ``*" and two belief sets $K$ and $K^\prime$ such
that $K^\prime \subseteq K$ and $K^\prime \setminus {K^\prime}^-_{\neg
A} = K \setminus K^-_{\neg A}$\footnote{\ldots, where the relationship
between ``*" and ``-" is given through the Levi Identity.}, the following
conditions hold:
\begin {enumerate}
\item [{\it a)}] 1. ${K^\prime}^-_{\neg A}\subseteq {K}^-_{\neg A}$;
({\it Monotony})\\
   2. ${K^\prime}^*_{\neg A}\subseteq {K}^*_{\neg A}$;
\item [{\it b)}] ${K^\prime}^-_{\neg A} \cup (K \setminus K^\prime)
\not\vdash {\neg A}$ if $\not\vdash {\neg A}$.
\end{enumerate}
\end{corollary}

In order to understand the significance of part {\it ii)} of Theorem 1,
consider the following situation: An agent makes inquiries in order to
attain a certain information-goal $G$ that would achieve its
desideratum.  In order to choose the right context, one should focus on
two aspects:
\begin {enumerate}
\item [I)] the context should include those sentences that help to
establish which beliefs are being retracted from the initial belief set
in order to accommodate the new evidence;
\item [II)] the context should include the beliefs that together with
the new one contribute to the derivation of the goal.
\end {enumerate}

If one is interested in choosing contexts that satisfy the first
requirement then the solution is to be found in Theorem 1 part {\it i)}.
Precisely, a minimal belief set $K^\prime_{m}$ is required. Note that
for a given $K$, the set of all $K^\prime$ satisfying the conditions (1)
and (2) specified by the theorem is partially ordered. Each totally
ordered maximal subset of this class has a minimal element, therefore
the class of $K^\prime$ has at least one minimal element $K^\prime_{m}$.

In order to meet requirement II) the context obeying condition I) must
satisfy additional restrictions, as specified in part {\it ii)} of
Theorem 1. Condition {\it ii)} stipulates the following: Supposing that
the goal can only be contextually derived and can be implied by neither
the new evidence $A$ nor the initial knowledge base $K$, we want to
choose such a context $K^\prime$ that does not become empty after
retracting $\neg A$ from it. This condition makes sense even when $\neg
A \notin K$ since it requires that $K^\prime$ is nonempty as long as the
goal is contextually derived from $A$. The second part of {\it ii)} says
that the goal will be obtained from the processing of the new evidence
$A$ in the context $K^\prime$. Also if the agent's original base
includes the negation of the goal, in order to derive the goal from the
revised knowledge base, $\neg G$ must be contained in the context $K^\prime$.
Note that when processing a piece of information $A$ in a context
$K^\prime$, if that belief is already part of the agent's old knowledge
base then there is no goal $G$ satisfying the conditions specified in
{\it ii)}. In this case, attaining goal $G$ is already achieved without
incorporating $A$ in $K$. The revision of $K^\prime$ with $A$ could only
result in a change in the entrenchment degree of $A$, provided that an
epistemic entrenchment ordering for $K$ (and for $K^\prime$) is assumed.

Belief sets that satisfy conditions (1) and ii) form a partially ordered
set having at least one $\subseteq$-minimal element. By choosing a minimal $K^\prime$, we
ensure the minimality requirement of an optimal context. The minimality
requirement is present also in the constructive approach to optimal
contexts.

Theorem 1 can be reformulated for bases of belief sets. The changes
consist of replacing belief set $K$ with the base $B_K$ and $K^\prime$
with a subset ${B^\prime}_K$ of $B_K$, $A \in K$ with $B_K \vdash A$,
etc.

\section{Contexts: Constructive Approach}
There are several ways to construct contexts: from entailment-sets of $\neg A$
and $G$, from maximal subsets of $K$ that fail to entail $\neg A$, using Grove's
system of spheres\footnote{See Grove (1988).} and from epistemic entrenchment
and accessibility
orderings. For lack of space, we will present only the first method
here. The others can be found in Vodislav (2000).

An {\it $A$-entailment set} is a set of sentences
each of which is essential for the derivation of the sentence $A$.

\begin{definition}
The set of sentences $X$ is {\em an $A$-entailment set in a belief set $K$} if and
only if: a) $X \subseteq K$; b) $A \in \mathit{Cn}$; c) for any proper subset $Y$ of $X$, $A \notin \mathit{Cn}(Y)$.\footnote{With some changes, the definition
is borrowed from Fuhrmann (1991), p.177.}
\end{definition}

Using the notion of ``entailment set of a sentence in a belief set" we
can define the context $K^\prime$ as the logical closure of the union of two
subsets, one responsible for the derivation of $\neg A$ and the other for the
entailment of $G$:
\begin{equation}
K^\prime = \mathit{Cn}[K^\prime_{\neg A} \cup K^\prime_G]
\end{equation}
where
\begin{itemize}
\item $K^\prime_{\neg A} = \bigcup (A\mbox{-entailment set in} K)$
\item $K^\prime_G = (\mathit{any}) X_G$ such that
$X_G \subseteq K$ and $X_G \cup \{A\}$ is a $G$-entailment set in $K^*_A$.
\end{itemize}

When constructing $K^\prime_G$, a construction for $K^*_A$ is assumed. For instance, $K^*_A$ can
be obtained from $K$ by means of the Levi Identity and partial meet
contraction functions.

The union $[K^\prime_{\neg A} \cup K^\prime_G]$ is not a belief set since entailment-sets for a sentence in a
belief set are not closed sets. However the closure of this union
(i.e. $K^\prime$) is both a belief set and a subset of $K$ since it is the closure of a
subset of $K$.

\begin{theorem}
Any subset $K^\prime \subseteq K$ that is generated from entailment-sets of $\neg A$ and $G$ as
stipulated by relation (3) satisfies Theorem 1.
\end{theorem}

\section{Accessibility Relation}
Theorem 1 describes contexts in a general manner without providing a
mechanism for defining a specific (optimal) context. In this section we
introduce the notion of accessibility as an additional structure for
belief sets that allows us to model optimal contexts.

This relation is defined over sentences in {\bf L}, as well as over
contexts/sets of sentences. The notion of accessibility is based on the
idea that depending on
the order in which our beliefs were acquired, we can assign them a
certain degree
of accessibility or we can rank them according to the rule: the last
acquired belief is the most accessible one.
We will use the notation $A \leq_a B$ for ``B is at
least as accessible as A".

The importance of the accessibility notion in the construction of
optimal contexts is supported by the ability of this concept to
characterize the effort required to access pieces of information from
someone's knowledge base. Further, using the accessibility notion one
can express the effort required to process some piece of information in
a certain context.

A set of sentences or a context will be considered as accessible as its
least accessible element.  The reason is that whenever one wants to
access a certain context or to process some piece of information in that
context, the effort required for such an operation is determined by the
effort necessary to access the least accessible sentence in that
context.

\begin {definition}
For any set of sentences $S$ in {\bf L}, if $A \in S$ and $A \leq_a B$
for all $B \in S$ then $S =_a A$.\footnote{$A=_a B$ iff $A \leq_a B$ and
$B \leq_a A$.}
\end{definition}

Instead of defining the accessibility of sentences in a quantitative
way, we rather axiomatically define the qualitative aspect of this
notion by introducing five new postulates.

\begin{definition}
Given a belief set $K$ in {\bf L}, an accessibility order related to $K$
is a relation $\leq_a$ on {\bf L} satisfying the following conditions:
\def\@listi{\leftmargin 25pt \labelwidth 20pt}
\begin{enumerate}
\item[{\rm A1)}]
If $A \leq_a B$ and $B \leq_a C$ then $A \leq_a C$; (transitivity)
\item[{\rm A2)}] For every two sentences $A,B \in \mathbf{L}$, either $A \leq_a B$ or
 $B \leq_a A$;
\item[{\rm A3)}] $A =_a \neg A$;
\item[{\rm A4)}] $A, \neg A \notin K$ if and only if $A \leq_a B$ for all $B \in \mathbf{L}$;
\item[{\rm A5)}] Given $B_K$ such that $\mathit{Cn}(B_K)=K$, for any set of sentences $X_A \subseteq B_K$ such that $X_A$ is the most
accessible $A$-entailment set in $B_K$ and $A \notin B_K$, the accessibility rank of $A$ is
determined by the accessibility rank of $X_A:A=_a X_A$.
\end{enumerate}
\end{definition}

The first two postulates define $\leq$ as a total transitive binary relation
over sentences in {\bf L}. Conditions (A3) and (A4) are better understood if
we consider that for a belief set $K$, the accessibility rank of a
sentence $A \in \mathbf{L}$ is determined by the lowest accessibility level to which
we need to go in $K$ in order to determine whether or not $A$ is the case
(given $K$).\footnote{By definition, we say that sentence $A$ {\em is the case
given} $K$ if and only
if $A \in K$; $A$ {\em is not the case given} $K$ if and only if $A \notin K$,
$\neg A \in K$,  and $A$ {\em is undetermined
given} $K$ if and only if $A,\neg A \notin K$.}
If $A \in K$, for instance, then the rank of $A$ is already known. If
$\neg A \in K$ then
$A$ has the same accessibility level as $\neg A$ since finding out that
$\neg A$ is the case leads
to finding out that $A$ is not the case. This explains postulate (A3). Also,
if neither $A$ nor $\neg A$ are in $K$
then it is impossible to find out given $K$ whether $A$ (and similarly,
$\neg A$) is the case or not.
Therefore, for this last case, $A$ would be among those sentences whose
level of accessibility is
the lowest. This gives us a reason for formulating the fourth postulate.

Note that condition (A5) does not hold in case $A$ is a
sentence in $B_K$. Since sentences in $B_K$ are independently acquired, they are
ranked according only to the order in which they have been incorporated
in the base, regardless of their logical relationship. In contrast, the
accessibility ranking of the information that is derived from the base
will be ranked only according to the ranks of the sentences from which
it is obtained by derivation.

Formally, we can describe the relation between the accessibility
ordering and optimal contexts as follows.

\begin{definition}
Consider some belief set $K$, a sentence $A \in \mathbf{L}$, and two
sub-theories of $K: K^\prime_1,K^\prime_2$,
both satisfying conditions (1) and ii) as specified in Theorem 1.
Then $K^\prime_1$ is preferred over $K^\prime_2$ as a potentially
optimal context if and only if
$K^\prime_1$ is more accessible than $K^\prime_2$.
\end{definition}

The postulates suggest the following conclusion and motivate
the next two definitions.

The number of accessibility levels in a base of a belief set is
the same as the number of accessibility levels in the belief set
generated from that base. Moreover, the accessibility ordering of the
base determines the accessibility ordering of the belief set generated
from the base. This property results from the last postulate, (A5).

\begin{definition}
For a belief set $K$, an accessibility ordering $\leq_a$ and a sentence
$A \in K$, we define $\mathit{cut}_{\leq_a}(A,K)=\{B \mid B \in K, A \leq_a B \}$.
\end{definition}

\begin{definition}
For a belief set $K$, a base $B_K$ of $K$, an accessibility ordering
$\leq_a$ and a sentence $A \in K$, we say that $\mathit{cut}_{\leq_a}(A,K)$ is a
{\it bad cut} if there is a sentence $C \in B_K$ such that $C \in
\mathit{Cn}[\mathit{cut}_{\leq_a}(A,K)]$ but $C \notin \mathit{cut}_{\leq_a}(A,K)$.
\end{definition}
Note then the following result:
\begin{theorem}
For an accessibility ordering $\leq_a$, a belief set $K$ and sentence $A
\in K$, if $\mathit{cut}_{\leq_a}(A,K)$ is not a bad cut then $\mathit{cut}_{\leq_a}(A,K)$
is a belief set.
\end{theorem}

This theorem provides a solution for the construction of optimal
contexts. Precisely, this result allows us to conclude that the subset
$K^\prime$ which contains $K$'s most accessible sentences and satisfies
conditions (1) and {\it ii)} as stipulated in Theorem 1 is or includes
an optimal context. However, if $K^\prime$ represents a bad cut, an
optimal context is a subset of the logical closure of $K^\prime$.

\section{Generating an Accessibility Ordering from a Finite Accessibility
Ranking}
In this section we define a finite accessibility ranking from which we
will then generate a finitely representable accessibility ordering. The
reason for this action is to overcome the representation problem
whenever an accessibility ordering ranks an infinite number of
sentences.  A finite accessibility ranking ranks the sentences of a
finite knowledge base. We will see that for axiomatizable
theories/belief sets we can generate an ordering that ranks all the
sentences of that theory/belief set from that finite ranking.

\begin{definition}
A finite accessibility ranking is an integer function $\mathit{af}:S\rightarrow (0,n]$ such
that $\mathit{range}(\mathit{af}) = [1,n]$; where $S$ is a finite set of sentences in {\bf L} and natural numbers 0
to $n$ designate ranks.
\end{definition}

The next definition introduces a way of extending the domain of the
finite ranking {\em af} to all the sentences in the language. The {\em degree of
accessibility} of a sentence in the language represents the effort
required to access someone's beliefs used to find out whether or not
that sentence is accepted by that person.

\begin{definition}
Let {\em af} be a finite accessibility ranking. The degree of
accessibility of a sentence $p \in \mathbf{L}$ is given as follows:
\begin{enumerate}
\item[a)] $\mathit{deg\_af}(p)=n$ if $\vdash p$;
\item[b)] $\mathit{deg\_af}(p)=\max\limits_i(\min\limits_q(\{\mathit{af}(q)\mid q\in X^i_p\}))$ if
$p\in \mathit{Cn}[\mathit{dom}(\mathit{af})]$ and $\not\vdash p$; where for each $i$, $X^i_p$ is a $p$-entailment set in
$\mathit{Cn}[\mathit{dom}(\mathit{af})]$
\item[c)] If $p \notin  \mathit{Cn}[\mathit{dom}(\mathit{af})]$, then $\mathit{deg\_af}(p)$ is determined inductively from the degree of
accessibility of sentences in $p \in \mathit{Cn}[\mathit{dom}(\mathit{af})]$ using the following rules:
\begin{enumerate}
\item[1)] $\mathit{deg\_af}(p)=\mathit{deg\_af}(\neg p)$;
\item[2)] $\mathit{deg\_af}(p)=0$ iff $p,\neg p \notin \mathit{Cn}[\mathit{dom}(\mathit{af})]$.
\end{enumerate}
\end{enumerate}
\end{definition}

The following theorem shows how a finite accessibility ranking can
generate an accessibility ordering using degrees of accessibility.

\begin{theorem}
Let {\it af} be a finite accessibility ranking and $p, q$ sentences
in {\bf L}. Define $\leq_\mathit{af}$ by $p \leq_\mathit{af} q$ iff
$\mathit{deg\_af}(p) \leq \mathit{deg\_af}(q)$. Then $\leq_\mathit{af}$
 is an accessibility ordering related to $\mathit{Cn}[\mathit{dom}(\mathit{af})]$.
\end{theorem}

Note that for $\mathit{dom}(\mathit{af})=B_K$, where $B_K$ is a finite set of sentences such that $\mathit{Cn}(B_K)=K$, $\leq_\mathit{af}$ is the
accessibility ordering related to $K$. Since $B_K$ is finite, $K$ is a finitely
axiomatizable theory.  In this case the finite accessibility ranking {\it af}
generates an accessibility ranking $\leq_\mathit{af}$ over $K$.

\section{Iterated Optimal Revision}
In the introduction it was mentioned that the inquirer's desideratum can
be fulfilled by several goals. Therefore, iterative revisions allow the
agent to get closer with each step to the fulfillment of its
desideratum. For each iteration a certain goal is attained. Each such
goal can either override the previously derived goal(s) or represent a
sub-goal of a final goal that is achieved after a number of such
iterations.

The adjustment that can be brought to the accessibility ranking of the
beliefs in a base of a belief set can be described as follows.

\begin{definition}
Let $B_K$ be a finite set of sentences in {\bf L}, $p \in B_K$ and {\it af} a finite
accessibility ranking assigned to $B_K$. We define the adjustment of {\it af},
when $B_K$ is revised with $A \in \mathbf{L}$, to be a function $\mathit{af}^*$ such that:\\
$\mathit{af}^*(p) = \left\{
\begin{array}{cl} n+1 & \mathrm{if } p=A \mathrm{\; or\; } p=\neg A \\
   a(p) & \mathrm{if } p \notin {B_K}^*_A \setminus B_K \\
  0 & \mathrm{otherwise}
\end{array}
\right.$
\end{definition}

This means that the newly acquired evidence is placed at the most
accessible level. All the beliefs in the original base are assigned
their initial level of accessibility. The sentences that are retracted
from $B_K$ when $A$ is incorporated are assigned the accessibility
ranking/degree as stipulated in part c) of Definition 8.

\section{Conclusions}
We have defined a way to change knowledge bases such that the same
effects as those achieved
in a conventional belief revision process are obtained, only with a
smaller processing effort.
This approach to belief revision is founded on notions such
as optimal context and accessibility. For the sentential model of belief
states we provide both a formal description of contexts as sub-theories
determined by three parameters and a method to construct contexts.
Next, we introduce an accessibility ordering for belief sets, which we
then use for selecting the best (optimal) contexts with respect to the
processing effort involved in the revision. Then, for finitely
axiomatizable knowledge bases, we characterize a finite accessibility
ranking from which the accessibility ordering for the entire base is
generated and show how to determine the ranking of an arbitrary sentence
in the language. Finally, we define the adjustment of the accessibility
ranking of a revised base of a belief set.

\section*{Acknowledgements}
This research was funded by NSERC Grant 0036853.
The referees' comments were appreciated.

\end{document}